\documentclass[letter, 10 pt, conference]{ieeeconf}  
\IEEEoverridecommandlockouts                              

\let\labelindent\relax
\usepackage[shortlabels,inline]{enumitem}
\usepackage{hyperref}
\usepackage{placeins}
\usepackage{amssymb}
\usepackage{pifont}
\usepackage{amsmath}
\usepackage{bm}
\usepackage{blindtext}
\usepackage{units}
\usepackage{multirow}
\usepackage[table]{xcolor}

\usepackage{tikz}
\usetikzlibrary{external,arrows,shapes,automata,backgrounds,petri,fit,arrows.meta,positioning,calc,matrix,chains,scopes}
\usepackage[skins]{tcolorbox}
\usepackage{pgfplotstable}
\usepackage{pgfplots}
\pgfplotsset{compat=newest}
\pgfplotsset{minor grid style={dashed,white!80!black}}
\pgfplotsset{major grid style={white!80!black}}

\definecolor{ownBlue}{rgb}{0.0705882,0.694118,0.839216}
\definecolor{ownGreen}{rgb}{0.631373,0.784314,0.25098}
\definecolor{ownGreen2}{rgb}{0.631373,0.884314,0.15098}
\definecolor{ownOrange}{rgb}{0.996078,0.741176,0.239216}
\definecolor{ownYellow}{rgb}{0.960784,0.898039,0.14902}
\definecolor{ownRed}{rgb}{0.990784,0.298039,0.14902}
\colorlet{ownRed2}{ownRed!80}

\usepackage{xstring}
\usepackage{collcell}
\usepackage{etoolbox}

\newcommand*{\MinNumber}{0.0}%
\newcommand*{\MidNumber}{0.5}%
\newcommand*{\MaxNumber}{1.0}%

\newcommand{\ApplyGradientI}[1]{%
	\iftoggle{inTableHeader}{%
		#1%
	}{
		\ifdim #1 pt > \MidNumber pt%
			\pgfmathsetmacro{\PercentColor}{max(min(100.0*(#1 - \MidNumber)/(\MaxNumber-\MidNumber),100.0),0.00)}%
			\edef\x{\noexpand\cellcolor{ownRed!\PercentColor!yellow}}%
			\x #1%
		\else%
			\pgfmathsetmacro{\PercentColor}{max(min(100.0*(\MidNumber - #1)/(\MidNumber-\MinNumber),100.0),0.00)}%
			\edef\x{\noexpand\cellcolor{ownGreen!\PercentColor!yellow}}%
			\x #1%
		\fi%
	}%
}

\newcommand{\ApplyGradientO}[1]{%
	\iftoggle{inTableHeader}{%
		#1%
	}{
		\ifdim #1 pt > \MidNumber pt%
			\pgfmathsetmacro{\PercentColor}{max(min(100.0*(#1 - \MidNumber)/(\MaxNumber-\MidNumber),100.0),0.00)}%
			\edef\x{\noexpand\cellcolor{ownGreen!\PercentColor!yellow}}%
			\x #1%
		\else%
			\ifdim #1 pt < \MinNumber pt%
				\edef\x{\noexpand\cellcolor{ownRed}}%
				\x #1%
			\else
				\pgfmathsetmacro{\PercentColor}{max(min(100.0*(\MidNumber - #1)/(\MidNumber-\MinNumber),100.0),0.00)}%
				\edef\x{\noexpand\cellcolor{ownRed!\PercentColor!yellow}}%
				\x #1%
			\fi
		\fi%
	}%
}

\newcommand{\SetGradient}[3]{%
	\renewcommand*{\MinNumber}{#1}%
	\renewcommand*{\MidNumber}{#2}%
	\renewcommand*{\MaxNumber}{#3}%
}

\newcolumntype{H}[3]{>{\SetGradient{#1}{#2}{#3}\collectcell\ApplyGradientO}c<{\endcollectcell}}
\newcolumntype{L}[3]{>{\SetGradient{#1}{#2}{#3}\collectcell\ApplyGradientI}c<{\endcollectcell}}

\newtoggle{inTableHeader}
\toggletrue{inTableHeader}
\newcommand*{\StartTableHeader}{\global\toggletrue{inTableHeader}}%
\newcommand*{\EndTableHeader}{\global\togglefalse{inTableHeader}}%
\let\OldTabular\tabular%
\let\OldEndTabular\endtabular%
\renewenvironment{tabular}{\StartTableHeader\OldTabular}{\OldEndTabular\StartTableHeader}%

\newcommand\copyrighttext{%
  \footnotesize \textcopyright 2022 IEEE. Personal use of this material is permitted. Permission from IEEE must be obtained for all other uses, in any current or future media, including reprinting/republishing this material for advertising or promotional purposes, creating new collective works, for resale or redistribution to servers or lists, or reuse of any copyrighted component of this work in other works.
  DOI: \href{https://ieeexplore-ieee-org.thi.idm.oclc.org/document/9827187}{10.1109/IV51971.2022.9827187}}
\newcommand\copyrightnotice{%
\begin{tikzpicture}[remember picture,overlay]
\node[anchor=south,yshift=10pt] at (current page.south) {\fbox{\parbox{\dimexpr\textwidth-\fboxsep-\fboxrule\relax}{\copyrighttext}}};
\end{tikzpicture}%
}

\title{\LARGE \bf 
Expert-LaSTS: \\Expert-Knowledge Guided Latent Space for Traffic Scenarios}

\author{Jonas Wurst$^{*}$, Lakshman Balasubramanian$^{*}$, Michael Botsch$^{*}$ and Wolfgang Utschick$^{+}$
\thanks{$^{*}$CARISSMA, Technische Hochschule Ingolstadt, 85049 Ingolstadt, Germany
        {\tt\small\{firstname.lastname\}@thi.de}}%
\thanks{$^{+}$Technical University of Munich, 80333 Munich, Germany        {\tt\small utschick@tum.de}}%
}

\begin{document}
\bstctlcite{IEEEexample:BSTcontrol}

\maketitle
\copyrightnotice
\thispagestyle{empty}
\pagestyle{empty}
\begin{abstract}
	Clustering traffic scenarios and detecting novel scenario types are required for scenario-based testing of autonomous vehicles. These tasks benefit from either good similarity measures or good representations for the traffic scenarios. In this work, an expert-knowledge aided representation learning for traffic scenarios is presented. The latent space so formed is used for successful clustering and novel scenario type detection. Expert-knowledge is used to define objectives that the latent representations of traffic scenarios shall fulfill. It is presented, how the network architecture and loss is designed from these objectives, thereby incorporating expert-knowledge. An automatic mining strategy for traffic scenarios is presented, such that no manual labeling is required. Results show the performance advantage compared to baseline methods. Additionally, extensive analysis of the latent space is performed.
\end{abstract}
\section{INTRODUCTION}\label{sec:intro}
Scenario-based testing is considered as one possible approach for the validation of \textit{Autonomous Vehicles} (AVs) \cite{Junietz2018a}. Two important tasks to enable the scenario-based approach are the definition of representative scenarios and the identification of potentially unknown and therefore untested scenarios. Representative scenarios can either be defined manually or automatically from collected data. For the latter one, usually clustering is used to define groups and representatives per group. The task of identifying untested scenarios can be realized through novelty detection (e.\,g. \cite{Wurst2021a,Langner2018a}) or by checking if the scenario fits into a group (e.\,g. \cite{Kruber2019b,Balasubramanian2021a}). For both tasks, clustering and novelty detection, a good representation or similarity measure is required. When applied directly on the plain data, the result is unsatisfactory (Sec. \ref{sec:COMP}). This work proposes a method to design a representation space for traffic scenarios using expert-knowledge constraints. This representation space can be utilized for the tasks of scenario clustering and of detecting novel scenario types. 

The method introduced in this work extends the findings of \cite{Wurst2021a}, where only the static part of a scenario is considered, i.\,e. the road infrastructure. The scenery is extended to include the ego dynamics as well. The ego dynamics are considered to be sufficient to represent a scenario in this work.

The methodology of this work can be summarized as follows. First, expert-knowledge based objectives are formulated. Then, it is shown how to design a loss function and network architecture such that those objectives are fulfilled. To realize the designed loss function, an automatic sample mining process is introduced, that is based on a similarity measure for scenarios. The paper proposes a similarity measure based on the topology graph of the road network and the routes defined by the trajectories. This way, no manual labeling is required. Overall, the objective is to form a latent space, which hierarchically divides samples into groups, based on infrastructure, route, and trajectory.\footnote{An implementation of the presented method can be found in \url{https://github.com/JWTHI/Expert-LaSTS}.}

The resulting latent space shows superior performance with respect to novel type detection, clustering and feature stability compared to alternative approaches. Possible applications of this work are in the field of validation for AVs. It can aid the analysis of existing scenario databases or the detection of novel scenario types.

The contributions of this work can be summarized as:
\begin{enumerate}
	\item Definition of an expert-knowledge aided loss and architecture to design the latent space as required.
	\item Definition of an automatic mining strategy for traffic scenarios.
	\item Comprehensive analysis and comparison of various representation spaces.
\end{enumerate}

The remaining is structured as follows. In Sec. \ref{sec:relWork}, related work is discussed. Then the method itself is presented in Sec. \ref{sec:method}. In Sec. \ref{sec:exps}, various representations are analyzed with respect to novel type detection, clustering and feature stability. The work is concluded, and an outlook is provided in Sec. \ref{sec:conc}.

\section{RELATED WORK}\label{sec:relWork}
Analyzing traffic scenarios for clustering and novel type detection has been the focus of many works recently. Either the analysis is performed based on an appropriate similarity measure or by generating representations of the scenarios.

\subsubsection{Similarity-Based}
In \cite{Kruber2018a} and the successor work \cite{Kruber2019b} a data-adaptive similarity measure is introduced based on the paths through an unsupervised random forest. This similarity measure  is used for clustering.

Finding scenario clusters is the objective of \cite{Langner2019a}. For this, abstract features are defined (e.\,g. Environment type, street type, curvature, average velocity, etc.) and collected in feature vectors, which are then clustered.

Another approach of using a specific similarity measure for clustering is presented in \cite{Kerber2020a}, where the average sum of the differences in the eight-car neighborhood is evaluated.

Clustering pairs of trajectories is the objective of \cite{Wang2020a}. Various approaches were examined, where the result showed that \textit{Dynamic Time Warping} (DTW) \cite{Sakoe1978a} in combination with $k$-means performed best in this study.

In \cite{Bernhard2021a} trajectories are clustered by hierarchical clustering. Each point of a trajectory is encoded as an area group label. An area group is defined by a Gaussian Mixture Model. The histograms of two encoded trajectories are compared via the Chi-squared distance.

Analyzing traffic scenarios from the ego information and other objects is realized in \cite{Ries2021a}. A procedure based on DTW and manual thresholds determines if a scenario is known.

Detecting unknown scenarios through novelty detection is the aim of \cite{Langner2018a}. There, an autoencoder is trained on known data. In test phase, if the reconstruction error is higher than a specific threshold, the scenario is assumed to be unknown.

In \cite{Wurst2020a}, novel traffic scenes based on infrastructure images are detected using a novel outlier detection method. The method utilizes local neighborhood similarities.

In contrast to this work the above stated works focus either on a specific similarity measure (e.\,g. DTW) or features, that are specifically selected to suit the used distance measure. Whereas in this work, the traffic scenarios are projected into a novel representation space.  

\subsubsection{Representation-Based}
The \textit{Principal Component Analysis} (PCA) is used for dimensionality reduction and hence to generate new representations  for clustering in \cite{Hauer2020a}. The PCA is applied to a column-wise normalized feature matrix, which is constructed using DTW on time series data.

In \cite{Harmening2020a} a deep learning network is used to reconstruct the input trajectories through a latent space. In the latent space, cluster analysis is performed.

Another approach utilizing deep learning is presented in \cite{Demetriou2020a}, where LSTMs \cite{Hochreiter1997a} are used to encode a trajectory and reconstruct it through a latent representation. The network is trained adversarial using a discriminator network as well. The latent representations are further projected with PCA and t-SNE \cite{Maaten2008a} before clustering. Moreover, the reconstruction error is used to estimate the novelty of a trajectory.

A different setting is presented in \cite{Balasubramanian2021b}, where some known classes and also unknown classes are assumed. In a multistep training process, a deep learning network is trained, such that it finds representations suited for classification and clustering of the unknown classes. The steps include self-supervised pre-training, classification and mixed training. For this, a novel representation based on the random forest is introduced. The results in the latent space are clustered. The training input consists of a sequence of images, representing the infrastructure and the objects as well.

Also in \cite{Zhao2021a}, an image sequence showing the infrastructure and the objects at each timestamp is used as input. Each frame is fed through an autoencoder which is trained using the reconstruction loss and a triplet loss. For the triplet loss, closer frames (time) shall be closer in the latent space. The latent representations of the image sequence are transformed into a sequence latent representation, which is then used for clustering. This transformation is realized by a recurrent neural network architecture, which aims to reconstruct frames and predict future frames. The sequence representations of the scenarios are clustered.

Instead of using deep learning, in \cite{Hoseini2021a} a tool chain of dimensionality reduction techniques and similarity measures is used for clustering. As input, a single trajectory is used.

This work builds on \cite{Wurst2021a}, where a method to project infrastructure images for novelty detection by utilizing expert-knowledge about the underlying topologies (c\,f. Sec. \ref{sec:method}).

The works summarized in this chapter generate an appropriate representation for traffic scenarios. Especially, \cite{Harmening2020a}, \cite{Demetriou2020a}, \cite{Balasubramanian2021b} and \cite{Zhao2021a} are comparable to this work since all of them use deep learning. However, the only two of them using a comparable input (infrastructure and dynamics) are \cite{Balasubramanian2021b} and \cite{Zhao2021a}. Since \cite{Balasubramanian2021b}, assumes known classes, it differs from this work. In contrast to \cite{Zhao2021a}, this work focuses on the ego dynamics. Moreover, expert-knowledge about the infrastructure and the trajectory is utilized.

\section{METHOD}\label{sec:method}
\begin{figure*}[t]
	\vspace{2mm}
	\centering
	\input{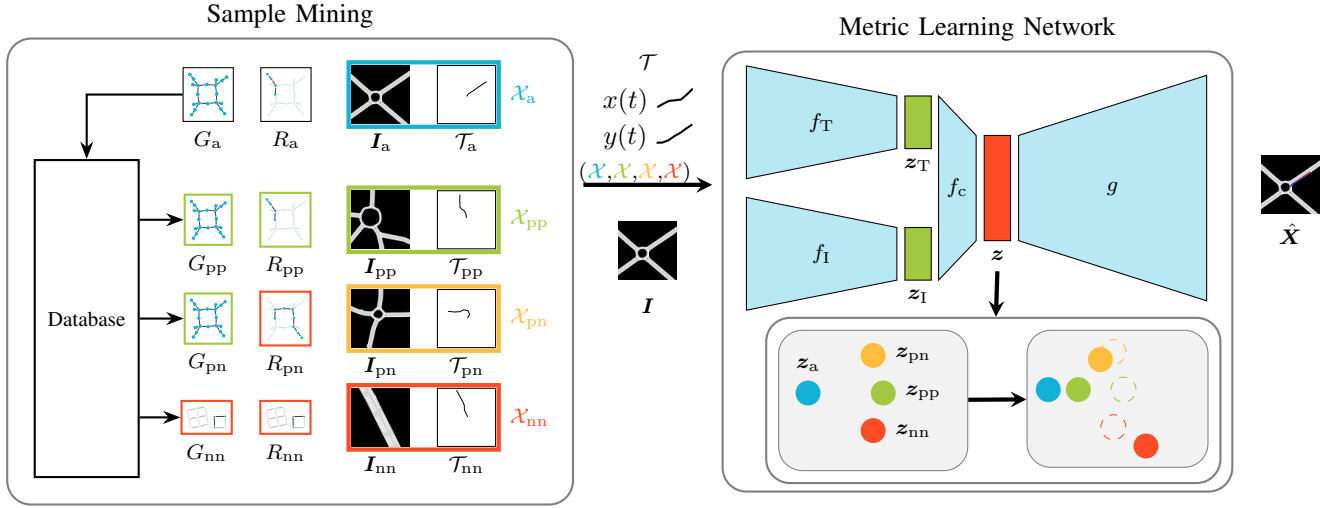}
	\caption{Metric Learning Network for Traffic Scenarios: The sample mining depicts how the required scenario quadruplet is selected based on the graphs $G$ and routes $R$. Each scenario $\mathcal{X}$ consists of an image $\bm{I}$ and a trajectory $\mathcal{T}$. Each scenario is processed by the network, leading to the latent representations $\bm{z}$. Below the network, the quadruplet learning objective is illustrated. The scenario is reconstructed through the decoder $g$ into a merged representation $\hat{\bm{X}}$}
	\label{fig:pipeline}
	\vspace{-0.5cm}
\end{figure*}
The aim of this work is to design a latent space by means of expert-knowledge to represent traffic scenarios. It is shown how this latent space can be utilized to detect unknown traffic scenario types and to cluster traffic scenarios. Here, a traffic scenario is described by the road infrastructure and the dynamic information of the ego. This work extends \cite{Wurst2021a}, which is limited to the infrastructure of a traffic scenario.

\subsection{Preliminaries}\label{sec:preliminaries}
A traffic scenario $\mathcal{X}=\left\lbrace\bm{I}, \mathcal{T}\right\rbrace$ consists of the road infrastructure image $\bm{I}$ and the ego trajectory $\mathcal{T}$. The dataset $\mathcal{D}=\left\lbrace(\mathcal{X}_0,G_0,R_0),\dots,(\mathcal{X}_M,G_M,R_M)\right\rbrace$ consists of $M$ scenarios, its elements are defined in the following.

\subsubsection{Infrastructure}
The road infrastructure is represented as a grayscale birds-eye view image $\bm{I} \in\mathbb{R}^{S\times S}$ and a graph representation $G$. The graph contains $N_\mathrm{G}$ lane pieces as vertices $V=\left\lbrace v_1,\dots,v_{N_\mathrm{G}}\right\rbrace$ and $N_\mathrm{E}$ edges $E=\left\lbrace e_1,\dots,e_{N_\mathrm{E}}\right\rbrace$ connecting them. The graph for a scene includes
\begin{enumerate} 
	\item all lanes, which are part of the ego route,
	\item all lanes up to and including the next intersection,
	\item all lanes of possible intersections in 1) and 2), and
	\item all lanes neighboring any lanes in 1) - 3).
\end{enumerate}
This way, the graph contains mainly the relevant lanes, whereas the image contains all lanes within the defined area.

\subsubsection{Trajectory}
The trajectory information is represented in two ways. The sequence based representation $\mathcal{T}=\left\lbrace[x_1,y_1,t_1],\dots,[x_N,y_N,t_N]\right\rbrace$ and the route representation $R=\left\lbrace r_1,\dots,r_{N_\mathrm{G}}\right\rbrace$. The construction of $R$ is realized as follows. For a trajectory point the corresponding vertices are $v(x_1,y_1)$, which leads to the vertices sequence for the trajectory as $\mathcal{T}_\mathrm{R}=\left\lbrace v(x_1,y_1),\dots,v(x_N,y_N)\right\rbrace$. The route representation $R$ is directly linked to $G$ as,
\begin{equation}
	r_n=\left\lbrace\begin{array}{ll}
		2 & \text{if } v_n \in v(x_1,y_1)\\
		1 & \text{if } v_n \in \mathcal{T}_\mathrm{R} \setminus v(x_1,y_1)\\
		0 & \text{else} \\
		\end{array}\right. .\label{eq:route_def}
\end{equation}
Hence, the vertex on which the trajectory starts is linked to $r_n=2$, all other vertices the trajectory passes lead to $r_n=1$.

\subsection{Base Method}
In \cite{Wurst2021a}, a triplet autoencoder is used to project the infrastructure image $\bm{I}$ to a latent representation. Using the triplet learning, the latent representations are optimized to represent the topology of the infrastructure as well as their shape. The main components of the base method can be split into the mining of the samples and the triplet learning. This subsection summarizes the base method with only infrastructures \cite{Wurst2021a}.

\subsubsection{Mining}
For triplet learning it is required to define a data triplet, what is realized in the so-called mining. In this application, a triplet consists of three infrastructure images $(\bm{I}_\mathrm{a},\bm{I}_\mathrm{p},\bm{I}_\mathrm{n})$. Given a random anchor image $\bm{I}_\mathrm{a}$, a positive example $\bm{I}_\mathrm{p}$ and negative example $\bm{I}_\mathrm{n}$ must be found. The positive example should be similar to the anchor, while the negative example should be dissimilar.

It is proposed to use the road topologies underlying each image to perform the triplet mining. The possible positive examples are all the images, which have the same graph as the anchor $G_\mathrm{a}$, while all possible negative examples are all other images. An example: if the anchor image shows a four-way roundabout, the positive example would also contain a four-way roundabout. However, the shape of the roundabouts are not necessarily the same. The negative example would show some other topology (e.\,g. intersection).

The similarity of two graphs is realized through isomorphism. $G_i$ and $G_j$ are isomorphic $G_i \cong G_j$, if there exists a bijection $p:V_i \rightarrow V_j$ such that $(u,v)\in E_i \iff (p(u),p(v))\in E_j$. This leads to the infrastructure similarity
\begin{equation}
	s_\mathrm{i}\left(G_i,G_j\right)=\left\lbrace\begin{array}{ll}
	1 & \text{if }G_i \cong G_j\\
	0 & \text{else}\\
	\end{array}\right. .\label{eq:infraSim}
\end{equation}

The positive example is drawn from the subset $\mathcal{G}_\mathrm{p}=\left\lbrace G \mid s_\mathrm{i}(G,G_\mathrm{a})=1\right\rbrace$, representing all samples which have the same topology as the anchor sample. Contrary, the negative example is drawn from the remaining samples $\mathcal{G}_\mathrm{n}=\left\lbrace G \mid s_\mathrm{i}(G,G_\mathrm{a})=0\right\rbrace$.

\subsubsection{Triplet Learning}
The network is a triplet autoencoder consisting of an encoder $f: \bm{I} \mapsto \bm{z}$ and a decoder $g: \bm{z} \mapsto \hat{\bm{I}}$, where $\bm{z}\in\mathbb{R}^L$ is the latent representation. Training the network is performed simultaneously by two approaches: the autoencoding regime and the triplet learning. This combination enables both, the topology based learning through the triplet strategy, and a low-level image similarity caused by the autoencoding objective.

While training, the samples of the triplet are passed through the encoder $(\bm{z}_\mathrm{a},\bm{z}_\mathrm{p},\bm{z}_\mathrm{n})$. These latent representations are used to determine the distances to the anchor $d_\mathrm{ap}=\vert\vert f(\bm{I}_\mathrm{a}),f(\bm{I}_\mathrm{p})\vert\vert_2^2$ and $d_\mathrm{an}=\vert\vert f(\bm{I}_\mathrm{a}),f(\bm{I}_\mathrm{n})\vert\vert_2^2$. The distances are used in the triplet loss \cite{Schroff2015a}, as
\begin{equation}\label{eq:tripletLoss}
	\mathcal{L}_\mathrm{tri}(\bm{I}_\mathrm{a},\bm{I}_\mathrm{p},\bm{I}_\mathrm{n}) = \max\left(\alpha + d_\mathrm{ap} - d_\mathrm{an},0\right).
\end{equation}
Minimizing the triplet loss is achieved by pushing the latent representation of the negative $\bm{z}_\mathrm{n}$ away and pulling the positive $\bm{z}_\mathrm{p}$ representation close to $\bm{z}_a$. This way, the triplet loss realizes the similarity as defined for the mining.

In order to ensure a high visual similarity between neighbors in the latent space, the autoencoder regime is adopted for the anchor sample. Therefore, the reconstruction loss
\begin{equation}
	\mathcal{L}_\mathrm{rec}(\bm{I}_\mathrm{a}) = \vert\vert \bm{I}_\mathrm{a} - g\left(f\left(\bm{I}_\mathrm{a}\right)\right)\vert\vert_2^2
\end{equation}
is used for the training as well. The loss to train the network is given as
\begin{equation}
	\mathcal{L} = \mathcal{L}_\mathrm{tri}+\mathcal{L}_\mathrm{rec}.
\end{equation}

\subsection{Proposed Method}
This work extends the static description from \cite{Wurst2021a} to a scenario by considering the dynamics of the ego vehicle. The network architecture is adjusted and the triplet learning is extended to quadruplet learning, called metric learning in the following. The overall concept is depicted in Fig. \ref{fig:pipeline}, which is divided into the mining process and the metric learning network.

The aim of this work is to design a latent space, which enables the clustering of scenarios and the detection of novel scenario types. Expert-knowledge is used to aid and constrain the training process. For this, the following objectives are formulated:
\begin{enumerate}[A)]
	\item Scenarios with the same infrastructure and similar trajectories shall be close together in the latent space.\label{case_a}
	\item Scenarios with the same infrastructure but different trajectories shall be close but not as close as \ref{case_a}.\label{case_b}
	\item Scenarios without the same infrastructure shall be farther away than \ref{case_b}.\label{case_c}
	\item The distance of scenarios according to \ref{case_a} shall be adjusted based on the similarity of the underlying actions.\label{case_a_sub}
	\item Neighbors should have high similarities with respect to trajectory and infrastructure features. \label{case_d}
\end{enumerate}
The objectives reflect expert-knowledge based assumptions about the similarity of scenarios in a hierarchical way. Hence, the latent space shall realize these expert-knowledge based hierarchical similarity objectives. In order to achieve the objectives, an automatic mining process, the metric learning as well as the network architecture are presented. 

\subsubsection{Mining}
In the base method it is shown how the identification of similar infrastructures can be realized through their topology. Hence, distinguishing the cases \ref{case_c} from \ref{case_a} or \ref{case_b} is possible. To realize the required separation between \ref{case_a} and \ref{case_b}, and therefore to include the trajectory information to the mining process, the mining definitions are extended.

Given the case that two graphs are isomorphic $G_i \cong G_j$, hence their infrastructure is the same, two trajectories are considered to be similar, if they share the same route within their graphs. The trajectories are transformed into the route representation $R$ which are directly linked to the respective graph (see Sec. \ref{sec:preliminaries}). Two scenarios share the same route if there exists a bijection $p$ on $G_i,G_j$ such that $R_i = p(R_j)$, which is formulated as $G_i \cong G_j \mid R_i = p(R_j)$. The route-based similarity measure is defined as
\begin{equation}
	s_\mathrm{r}\left(G_i,G_j\right)=\left\lbrace\begin{array}{ll}
	1 & \text{if }G_i \cong G_j \mid R_i = p(R_j)\\
	0 & \text{else}\\
	\end{array}\right. .\label{eq:routeSim}
\end{equation}

According to \ref{case_a_sub}, just defining two trajectories to be similar is not sufficient, instead it shall be adjusted based on the actions of the trajectories. To allow further fine-tuning in the training, for trajectories sharing a similar route an additional similarity measure $s_\mathrm{t}$ is defined. Let $\mathcal{A}_i = [\bm{a}_\mathrm{lat},\bm{a}_\mathrm{lon},\vert v\vert]$ be the accelerations and speed per timestamp for the $i$th trajectory. The dissimilarity between two trajectories is then calculated via $d = d_\mathrm{DTW}\left(\mathcal{A}_i,\mathcal{A}_j\right) \vert\mathrm{DTW}_\mathrm{seq}\vert$, where $d_\mathrm{DTW}$ denotes the DTW distance and $\vert\mathrm{DTW}_\mathrm{seq}\vert$ the warping path length divided by maximum sequence length. The intuition is to compare the trajectories which share the same route on an action level, therefore considering the accelerations and speed. The similarity is calculated with respect to maximum dissimilarity ($d_{\mathrm{max}}$) within all trajectories with the same route, as 
\begin{equation}
	s_\mathrm{t} = 1- \frac{d}{d_{\mathrm{max}}}.
\end{equation}

The mining of a data quadruplet is realized by randomly sampling an anchor scenario $\mathcal{X}_\mathrm{a}$. Based on its graph $G_\mathrm{a}$ and route $R_\mathrm{a}$, samples for the cases \ref{case_a} - \ref{case_c} are drawn. The example scenario having similar infrastructure and similar route $\mathcal{X}_\mathrm{pp}$ is sampled based on $\mathcal{G}_\mathrm{pp}=\left\lbrace G \mid s_\mathrm{i}(G,G_\mathrm{a})=1,s_\mathrm{r}(G,G_\mathrm{a})=1\right\rbrace$. For sampling the scenario with same infrastructure but different route $\mathcal{X}_\mathrm{pn}$, $\mathcal{G}_\mathrm{pn}=\left\lbrace G \mid s_\mathrm{i}(G,G_\mathrm{a})=1,s_\mathrm{r}(G,G_\mathrm{a})=0\right\rbrace$ is used. Finally, the scenario with a different infrastructure $\mathcal{X}_\mathrm{nn}$ is sampled from $\mathcal{G}_\mathrm{nn}=\left\lbrace G \mid s_\mathrm{i}(G,G_\mathrm{a})=0\right\rbrace$. This mining process is also visualized on the left side of Fig. \ref{fig:pipeline}.

\subsubsection{Metric Learning}
This section introduces the metric learning and the network, such that the objectives \ref{case_a} - \ref{case_d} are realized. 

As shown in Fig. \ref{fig:pipeline}, the network consists of two encoders, $f_\mathrm{I}: \bm{I} \mapsto \bm{z}_\mathrm{I}\in\mathbb{R}^{L_\mathrm{I}}$ for the image and $f_\mathrm{T}: \mathcal{T} \mapsto \bm{z}_\mathrm{T}\in\mathbb{R}^{L_\mathrm{T}}$ for the trajectory. The two intermediate representations are concatenated and then passed through the network $f_\mathrm{c}: [\bm{z}_\mathrm{T},\bm{z}_\mathrm{I}]\mapsto \bm{z}$ to create the final latent representation $\bm{z}\in \mathbb{R}^L$. Finally, a decoder $g: \bm{z} \mapsto \hat{\bm{X}}$ is used to generate a merged representation $\hat{\bm{X}}\in \mathbb{R}^{S \times S \times 2}$ of infrastructure and trajectory.

The data quadruplet is passed through the encoder, such that the latent representations $\bm{z}_\mathrm{a},\bm{z}_\mathrm{pp},\bm{z}_\mathrm{pn},\bm{z}_\mathrm{nn}$ are generated. The squared distances from the anchor to the examples are defined as  $d_\mathrm{pp} = \Vert \bm{z}_\mathrm{a}-\bm{z}_\mathrm{pp} \Vert^2_2$, $d_\mathrm{pn} = \Vert \bm{z}_\mathrm{a}-\bm{z}_\mathrm{pn} \Vert^2_2$ and $d_\mathrm{nn} = \Vert \bm{z}_\mathrm{a}-\bm{z}_\mathrm{nn} \Vert^2_2$. The objectives \ref{case_a} - \ref{case_a_sub} can then be formulated as
\begin{align}
	d_\mathrm{nn} &\geq d_\mathrm{pn} + \alpha_\mathrm{G},\\
	d_\mathrm{pn} &\geq \max\left\lbrace d_\mathrm{pp},\alpha_\mathrm{T}\right\rbrace + \alpha_\mathrm{R},\\
	d_\mathrm{pp} & = (1-s_\mathrm{t})\alpha_\mathrm{T},
\end{align}
where $\alpha_{\dots}$ are margin parameters. Those constraints lead to the following loss formulations
\begin{align}
	\mathcal{L}_\mathrm{G} &= \max\left\lbrace \alpha_\mathrm{G} + d_\mathrm{pn} - d_\mathrm{nn} ,0\right\rbrace,\label{eq:graphLoss}\\
	\mathcal{L}_\mathrm{R} &= \max\left\lbrace \alpha_\mathrm{R} + \max\left\lbrace\alpha_\mathrm{T},d_\mathrm{pp}\right\rbrace -d_\mathrm{pn} ,0\right\rbrace,\label{eq:routeLoss}\\
	\mathcal{L}_\mathrm{T} &= \vert(1-s)\alpha_\mathrm{T} - d_\mathrm{pp}\vert.\label{eq:trajLoss}
\end{align}
The losses Eq. \ref{eq:graphLoss} and Eq. \ref{eq:routeLoss} are basic triplet losses \cite{Schroff2015a}, when optimizing both, it leads to the quadruplet loss as presented in \cite{Zhang2016a}.

The objective in \ref{case_d} is not directly addressed by the former loss definitions, hence another strategy needs to be adopted. For this purpose, the latent representation of a scenario is used to reconstruct the scenario. In this case, the reconstruction generates an image with two channels, one channel for the infrastructure as in $\bm{I}$ and one channel for the trajectory information as in $\mathcal{T}$. This way, the network has to connect the image information with the trajectory information. Furthermore, for the decoding to work properly, the neighborhood in the latent space has to share high similarities. To further aid the training, all infrastructure parts, that are not part of the graph are removed from the target reconstruction image. Since simple reconstruction might fail due to the high sparsity of the generated output, the reconstruction loss is adopted piece-wise for trajectory, infrastructure, and respective background pixels. The weighted sparse reconstruction loss is defined as
\begin{equation}
	\mathcal{L}_\mathrm{Rec} = \gamma_{\mathrm{I}} \mathcal{R}(\mathcal{I}_\mathrm{I}) + \gamma_{\bar{\mathrm{I}}}\mathcal{R}(\mathcal{I}_{\bar{\mathrm{I}}})+ \gamma_{\mathrm{T}} \mathcal{R}(\mathcal{I}_\mathrm{T}) + \gamma_{\bar{\mathrm{T}}}\mathcal{R}(\mathcal{I}_{\bar{\mathrm{T}}}),
\end{equation}
with $\mathcal{R}(\mathcal{I})= \frac{1}{\vert\mathcal{I}\vert}\sum_{i\in\mathcal{I}} (\bm{X}_\mathrm{a}(i)-\hat{\bm{X}}_\mathrm{a}(i))^2$ the reconstruction error for the pixel subset $\mathcal{I}$. The subset $\mathcal{I}_\mathrm{I}$ refers to all ground truth infrastructure pixels, $\mathcal{I}_{\bar{\mathrm{I}}}$ all remaining pixels in the infrastructure channel. For the trajectory pixel sets $\mathcal{I}_\mathrm{T}$ and $\mathcal{I}_{\bar{\mathrm{T}}}$ the logic applies respectively.

Combining all the loss definitions leads the overall loss as
\begin{equation}
	\mathcal{L} = \beta_\mathrm{M}\left(\beta_\mathrm{G}\mathcal{L}_\mathrm{G} + \beta_\mathrm{R}\mathcal{L}_\mathrm{R} + \beta_\mathrm{T}\mathcal{L}_\mathrm{T}\right) + \beta_\mathrm{Rec}\mathcal{L}_\mathrm{Rec}.
\end{equation}
Training the network with $\mathcal{L}$ aims to realize the expert objectives as defined in the beginning of this section.
\FloatBarrier

\section{EXPERIMENTS}\label{sec:exps}
The quality of the latent space constructed by the proposed method is analyzed through various experiments. The analysis is based on four perspectives: 
\begin{enumerate*}
	\item Novel type detection: \textit{Given a base set, are new scenario types detected as novel?}
	\item Clustering: \textit{Do the latent representations form meaningful clusters?}
	\item Feature stability: \textit{Are scenario features of neighboring latent representations similar?}
	\item Visualization: \textit{Can the latent space be analyzed through aided visualizations?}
\end{enumerate*}
In order to evaluate the impact of the various loss terms and possible network variations, different settings are used for all the experiments. 

The section is structured as follows. First, the dataset is explained. Second, the different network settings and training variants are summarized. The novel scenario type detection analysis is shown in the third part, followed by the clustering analysis in fourth and the feature stability in the fifth part. The visual assessment is shown in part six. The different results are summarized in the last part.

\subsection{Data}
The data used to train and analyze the network is generated through simulation. The data generation process is divided in two parts, the infrastructure sampling and the simulation.

\subsubsection{Infrastructure Sampling}
Sampling the infrastructures is realized as in \cite{Wurst2021a}. From OpenStreetMap (OSM) \cite{OpenStreetMap2021}, random nodes are selected as center for the scenarios. The underlying road infrastructure within the area of $\unit[200]{m}\times\unit[200]{m}$ per node is extracted as image ($\unit[100]{px}\times\unit[100]{px}$), graph and as map for simulation. To generate the graphs and images, the tools from \cite{Wurst2021a} are used. This way $\approx 70\,000$ infrastructures are extracted.

\subsubsection{Simulation}
For each of the infrastructures, simulations in SUMO \cite{SUMO2018} are performed. One vehicle (ego) is inserted at the center position of the scenario, while other vehicles are randomly spawned, such that the scenarios show a rather high traffic load. The scenario is simulated for a total span of \unit[6]{s}. The ego information during this timespan is recorded.

The infrastructures as well as the routes are sampled such, that for each scenario, similar infrastructure and route examples are available.

\subsubsection{Groups}\label{sec:groups}
For the analysis with respect to groups (clustering and novelty detection), three detail levels are examined. First, a rough level, consisting of the categories: \begin{enumerate*}
	\item single-lane,
	\item multi-lane,
	\item intersection,
	\item intersection entering,
	\item roundabout,
	\item roundabout entering,
	\item highway and 
	\item highway entering.
\end{enumerate*}
This leads to the subscript $\dots_\mathrm{C}$.

The second detail level considers all unique infrastructure graphs in the dataset, leading to $704$ groups. Hence, those groups can be used to analyze the performance with respect to the infrastructure. The subscript $\dots_\mathrm{G}$ is used to indicate the usage of the second level.

The third and most detailed level is provided by the $2330$ groups, formed by all unique routes combined with their graphs. It provides insight with respect to the complete scenario. Here, the subscript $\dots_\mathrm{R}$ is used.

\subsection{Comparison}\label{sec:COMP}
\begin{table*}[htbp]
	\vspace{2mm}
	\renewcommand{\arraystretch}{1.3}
	\centering\begin{tabular}{c||H{0.543}{0.99}{0.999}|H{0.628}{0.919}{0.943}|H{0.617}{0.904}{0.912}|||H{0.319}{0.839}{0.996}|H{0.218}{0.90}{0.910}|H{0.167}{0.622}{0.622}|||L{31}{36.67}{39.99}||L{0.54}{0.58}{1.37}|L{1.43}{1.91}{5.05}|L{0.51}{0.57}{0.77}|L{0.36}{0.37}{0.43}|L{0.18}{0.20}{0.34}}
		&\multicolumn{3}{c|||}{Novelty Detection \ref{sec:noveltyC}} & \multicolumn{3}{c|||}{Clustering \ref{sec:clusteringC}}& \multicolumn{6}{c}{Feature Stability \ref{sec:featsC}}\\\cline{2-13}
		Approach &${AUC}_\mathrm{C}$ & ${AUC}_\mathrm{G}$ & ${AUC}_\mathrm{R}$& ${ACC}_\mathrm{C}$ & ${ACC}_\mathrm{G}$ & ${ACC}_\mathrm{R}$ & $\bar{d}_I$ & $\bar{d}_T$ & $\bar{d}_v$ & $\displaystyle \bar{d}_{a_\text{lon}}$ & $\displaystyle \bar{d}_{a_\text{lat}}$ & $\bar{d}_{\psi}$\\
		\hline\hline
		\textbf{Proposed}							& 0.991 & 0.919 & 0.904 & 0.839 & 0.900 & 0.622	& 36.67 & 0.58 	& 1.91 	& 0.57 	& 0.37 	& 0.20\EndTableHeader\\\hline
		\texttt{plain}								& 0.500	& 0.485	& 0.487 & 0.265 & 0.255 & 0.271 & 28.80	& 1.47	& 5.96	& 0.74	& 0.43	& 0.30\\\hline
		\texttt{UMAP}								& 0.654 & 0.557	& 0.574 & 0.406 & 0.070	& 0.070 & 36.57	& 1.37	& 5.56	& 0.84	& 0.44	& 0.35\\\hline
		\texttt{PCA}								& 0.543	& 0.632	& 0.617	& 0.265 & 0.342	& 0.324 & 30.89	& 1.27	& 5.09	& 0.77	& 0.44	& 0.32\\\hline
		\texttt{Classifier}							& 0.975	& 0.885	& 0.860	& 0.319	& 0.302	& 0.348	& 40.61	& 1.01 	& 2.89	& 0.63	& 0.43	& 0.34\\\hline
		\texttt{Autoencoder}						& 0.786 & 0.628 & 0.641 & 0.503 & 0.218 & 0.167	& 36.53 & 0.54 	& 2.02 	& 0.65 	& 0.37 	& 0.18\\\hline
	\end{tabular}
	\caption{Comparison Performance Summary: red indicates worse than the proposed method, green better and yellow comparable.}
	\label{tb:resultsComp}
	\vspace{-1.0cm}
\end{table*}
\subsubsection{Alternative Approaches}
As mentioned in Sec. \ref{sec:relWork}, typical approaches either operate in the input space or perform dimensionality reduction (t.SNE etc.). The proposed method is compared to the following alternatives.
\paragraph{\texttt{Plain}} The input is used as representation directly. Hence, the image and the trajectory are vectorized and concatenated (10090 dimensions).
\paragraph{\texttt{UMAP}} The \texttt{plain} input is projected with UMAP \cite{McInnes2018a} to 64-dimensional space.
\paragraph{\texttt{PCA}} The first 64 principal components (PCA) of the \texttt{plain} input are used.
\paragraph{\texttt{Classifier}} The latent representations when using the same architecture as the proposed method but replacing the decoder with two classification heads, one for the 704 unique graphs (64-704-704-704) and one for the 2330 unique routes (64-2330-2330-2330). The used representation is the intermediate 64-dimenional latent representation, not the classification output.
\paragraph{\texttt{Autoencoder}} The latent representations when using the same architecture as the proposed method but without metric learning just operating as autoencoder.

\subsubsection{Novel Scenario Type Detection}\label{sec:noveltyC}
Detecting novel scenario types is a crucial task in the validation process of autonomous driving. The latent space designed by the former method suits this need, as shown in this section. In this work, detecting novel scenario types is realized through outlier detection. Therefore, assuming a base data set (the already known scenarios), the task is to identify scenarios which do not fit in the base data set.

The novelty detection is performed as n-vs-1, where one group (Sec. \ref{sec:groups}) is excluded from the base dataset. It is tested, how well this left-out group is detected as novel. This procedure is repeated for all groups. The novelty detection performance is measured using the \textit{Area Under Curve} (AUC). \textit{Angle Based Outlier Detection} (ABOD) is used as the novelty detection method.

As the results show (Tb. \ref{tb:resultsComp}), detecting novel scenario types is best realized in the latent space formed by the proposed method. The only other method reaching considerable performance is the classifier based approach. The alternative approaches miss out noticeably when compared to the proposed method.

\subsubsection{Clustering}\label{sec:clusteringC}
Another task in the field of validating AVs is to cluster scenarios into groups. This way, possible representatives for testing per cluster can be defined. In this subsection, the clustering performance when using the designed latent space is demonstrated.

Because of the highly imbalanced number of samples per group (Sec. \ref{sec:groups}), agglomerative clustering suits this task well. As linkage function, average is used. The clustering performance is stated as accuracy ${ACC}_{\dots}$ \cite{Yang2010a}. For this, the best mapping between the ground truth labels and the predicted labels is determined. Given this mapping the accuracy can be determined.

For the clustering, none of the alternative approaches reached comparable performance to the proposed method (Tb. \ref{tb:resultsComp}). Even the classifier based model is not able to provide sufficient clustering results. This clearly shows the necessity for the expert-knowledge designed latent space for clustering traffic scenarios.

\subsubsection{Feature Stability}\label{sec:featsC}
As stated in the design requirements \ref{case_d}, one of the objectives is that neighbors in the latent space share high similarities with respect to various features. An analysis accessing this is realized in this section.

To analyze the stability within a neighborhood, for each data point, the 15 nearest neighbors in the latent space are considered for the further analysis. The average differences from the data points in focus to their neighbors are determined. For calculating the differences $d_{\dots}$ various features are used: \begin{enumerate*}
	\item $d_I$ image difference (like in \cite{Wurst2021a}),
	\item $d_T$ trajectory difference (average displacement),
	\item $d_v$ average velocity difference,\label{feat:newstart}
	\item $\displaystyle d_{a_\mathrm{lon}}$ average longitudinal acceleration difference,
	\item $\displaystyle d_{a_\mathrm{lat}}$ average lateral acceleration difference and
	\item $d_\psi$ average orientation difference.\label{feat:newend}
\end{enumerate*} 
Those values are averaged over the complete data set, leading to $\bar{d}_{\dots}$. The smaller those average values, the more similar the features within the neighborhood, hence the better the objective is fulfilled. It is important to note, that the features \ref{feat:newstart} - \ref{feat:newend} are not part of the input.

The results are listed in Tb. \ref{tb:resultsComp} column feature stability. The performance for the difference of images is better for the most of the alternative approaches except for the classifier. When comparing the trajectory features, however, the proposed model outperforms all alternative approaches except the autoencoder. If neighboring data points shall share high similarities with respect to features from both domains (infrastructure and trajectory), the latent spaces provided by the proposed method or the autoencoder are the best choices.

\subsection{Ablation}
\begin{table*}[htbp]
	\vspace{2mm}
	\renewcommand{\arraystretch}{1.3}
	\centering\begin{tabular}{c||H{0.98}{0.99}{0.999}|H{0.88}{0.919}{0.943}|H{0.880}{0.904}{0.912}|||H{0.76}{0.839}{0.996}|H{0.7}{0.90}{0.910}|H{0.5}{0.622}{0.622}|||L{36.21}{36.67}{36.99}||L{0.54}{0.58}{0.60}|L{1.43}{1.91}{2.02}|L{0.51}{0.57}{0.61}|L{0.36}{0.37}{0.41}|L{0.18}{0.20}{0.25}}
		&\multicolumn{3}{c|||}{Novelty Detection \ref{sec:noveltyAB}} & \multicolumn{3}{c|||}{Clustering \ref{sec:clusteringAB}}& \multicolumn{6}{c}{Feature Stability \ref{sec:featsAB}}\\\cline{2-13}
		Setting &${AUC}_\mathrm{C}$ & ${AUC}_\mathrm{G}$ & ${AUC}_\mathrm{R}$& ${ACC}_\mathrm{C}$ & ${ACC}_\mathrm{G}$ & ${ACC}_\mathrm{R}$ & $\bar{d}_I$ & $\bar{d}_T$ & $\bar{d}_v$ & $\displaystyle \bar{d}_{a_\text{lon}}$ & $\displaystyle \bar{d}_{a_\text{lat}}$ & $\bar{d}_{\psi}$\\
		\hline\hline
		\textbf{Proposed}							& 0.991 & 0.919 & 0.904 & 0.839 & 0.900 & 0.622	& 36.67 & 0.58 	& 1.91 	& 0.57 	& 0.37 	& 0.20\EndTableHeader\\\hline
		$\beta_\mathrm{T}=0$ 						& 0.986 & 0.904 & 0.879 & 0.948 & 0.880 & 0.507	& 36.59 & 0.63 	& 2.30 	& 0.61 	& 0.37 	& 0.20\\\hline
		$\beta_\mathrm{T}=0$, $\beta_\mathrm{R}=0$  & 0.987 & 0.883 & 0.847 & 0.757 & 0.753 & 0.379	& 36.99 & 0.57 	& 2.05 	& 0.59 	& 0.37 	& 0.19\\\hline
		$\beta_\mathrm{M}=0$ 						& 0.786 & 0.628 & 0.641 & 0.503 & 0.218 & 0.167	& 36.53 & 0.54 	& 2.02 	& 0.65 	& 0.37 	& 0.18\\\hline
		$\beta_\mathrm{Rec}=0$ 						& 0.980 & 0.943 & 0.912 & 0.712 & 0.917 & 0.541	& 39.91 & 0.77 	& 1.43 	& 0.51 	& 0.41 	& 0.31\\\hline
		$\alpha_\mathrm{G}=10$, $\alpha_\mathrm{R}=5$& 0.990	& 0.919	& 0.898	& 0.705 & 0.857 & 0.383 & 37.66 & 0.64  & 2.19  & 0.57  & 0.38  & 0.21\\\hline
		$L_{\dots}=L_{\dots}*2$ 					& 0.991 & 0.915 & 0.904 & 0.611 & 0.910 & 0.595	& 36.21 & 0.55 	& 1.80 	& 0.55 	& 0.36 	& 0.19\\\hline
		$f_\mathrm{I}:$ ViT  						& 0.965 & 0.883 & 0.880 & 0.768 & 0.722 & 0.620	& 36.70 & 0.59 	& 2.05 	& 0.56 	& 0.37 	& 0.20\\\hline
		$f_\mathrm{T}:$ LSTM  						& 0.991 & 0.919 & 0.904 & 0.848 & 0.906 & 0.605	& 36.54 & 0.62 	& 2.23 	& 0.61 	& 0.37 	& 0.21\\\hline
		\texttt{random-excl}						& 0.999 & 0.866 & 0.855 & 0.996 & 0.611 & 0.460	& 36.12 & 0.56 	& 1.94 	& 0.59 	& 0.36 	& 0.19\\\hline
		\texttt{group}								& 0.821 & 0.873 & 0.871 & 0.344 & 0.542 & 0.520	& 36.34 & 0.58 	& 1.94 	& 0.61 	& 0.37 	& 0.20\\\hline
	\end{tabular}
	\caption{Ablation Performance Summary: red indicates worse than the proposed method, green better and yellow comparable.}
	\label{tb:resultsAB}
	\vspace{-1.0cm}
\end{table*}
\subsubsection{Model Variants}
To assess the impact of the various possible settings of the network and the learning, they are varied and compared.

\textbf{Proposed Setting:} The setting which shows overall good perfomance is as follows: $f_\mathrm{I}$: ResNet-18 \cite{He2016a}, $f_\mathrm{T}$: Transformer-Encoder \cite{Vaswani2017a}, $L_\mathrm{I} =64$, $L_\mathrm{T} =16$, $L =64$, $\beta_\mathrm{M}=\beta_\mathrm{G}=\beta_\mathrm{R}=\beta_\mathrm{T}=1$, $\beta_\mathrm{Rec}=10$, $\gamma_\mathrm{I}=\gamma_\mathrm{T}=5$, $\gamma_\mathrm{\bar{I}}= 10$, $\gamma_\mathrm{\bar{T}}=20$, $\alpha_\mathrm{G}=\alpha_\mathrm{R}=\alpha_\mathrm{T}=1$ and \texttt{random} negative sampling.

In the Transformer-Encoder ($f_\mathrm{T}$), an embedding token is used like in \cite{Wurst2021a}. As alternative $f_\mathrm{T}$, a LSTM \cite{Hochreiter1997a} is used. And as alternative image encoder $f_\mathrm{I}$, a ViT \cite{Dosovitskiy2021a} with patch-size of 10, dimensionality of 256, MLP dimensionality of 128, 16 layers and 16 heads is used.

All the other variants used in the ablation, adjust few parameters from the proposed setting. For example, in the variant $\beta_\mathrm{M}=0$ just the according value is changed, all other values are as stated above.

For the negative sampling, the following strategies are examined. \texttt{random}: from all graphs that are different, \texttt{group}: from all graphs that are different but only inside the same category (highway, etc.) and \texttt{random-excl}: from all graphs that are different excluding the same category.
\subsubsection{Novel Scenario Type Detection}\label{sec:noveltyAB}
In Tb. \ref{tb:resultsAB}, the results for various model variations are shown. The description in the left column, indicates what parameters are changed compared to the proposed setting.

As one can see, detecting novel scenario types is realized best either with the proposed setting, $\alpha_\mathrm{G}=10$ and $\alpha_\mathrm{R}=5$, $L_{\dots}=L_{\dots}*2$ or $f_\mathrm{T}:$ LSTM settings. Hence, the choice of the margin parameters has neglectable effect on the performance. Also, the size of the latent space size is rather irrelevant in terms of novelty detection. The same holds for the selected trajectory encoder.

The importance of each loss terms can be seen from\\ Tb. \ref{tb:resultsAB} (rows 2-5). The metric learning related losses ($\mathcal{L}_\mathrm{M}$,  $\mathcal{L}_\mathrm{G}$, $\mathcal{L}_\mathrm{R}$ and $\mathcal{L}_\mathrm{T}$) are required for good performance.

\subsubsection{Clustering}\label{sec:clusteringAB}
The clustering accuracies for the model variants are shown in the corresponding columns in Tb. \ref{tb:resultsAB}. Only the model variant when using LSTM encoder is achieving comparable results to the proposed setting.

As for the novelty detection, the metric learning related losses ($\mathcal{L}_\mathrm{M}$,  $\mathcal{L}_\mathrm{G}$, $\mathcal{L}_\mathrm{R}$ and $\mathcal{L}_\mathrm{T}$) are important.

\subsubsection{Feature Stability}\label{sec:featsAB}
The double sized latent space setting $L_{\dots}=L_{\dots}*2$ achieves better results than the proposed setting. Changing the negative sampling strategy to \texttt{random-excl} does only slightly affect the performance in terms of feature stability. 

Using only the reconstruction loss ($\beta_\mathrm{M}=0$) does not outperform the proposed setting in terms of feature stability. Hence, the expert-knowledge aided losses ($\mathcal{L}_\mathrm{G}$, $\mathcal{L}_\mathrm{R}$, $\mathcal{L}_\mathrm{T}$) help in structuring the latent space also for the stability criterion. Not using the reconstruction loss ($\beta_\mathrm{Rec}=0$) has a negative effect for the most features. This supports the designed intuition to use the autoencoder regime to achieve feature stability.

\subsection{Visualization}
\setlength{\unitlength}{1ex}
\begin{figure}[h]
	\vspace{-4mm}
	\centering
	\setlength\tabcolsep{1pt}
	\begin{tabular}{c c c}
		Categories & $\bar{v}$ & $\bar{\psi}$\\
		\includegraphics[width=0.33\columnwidth]{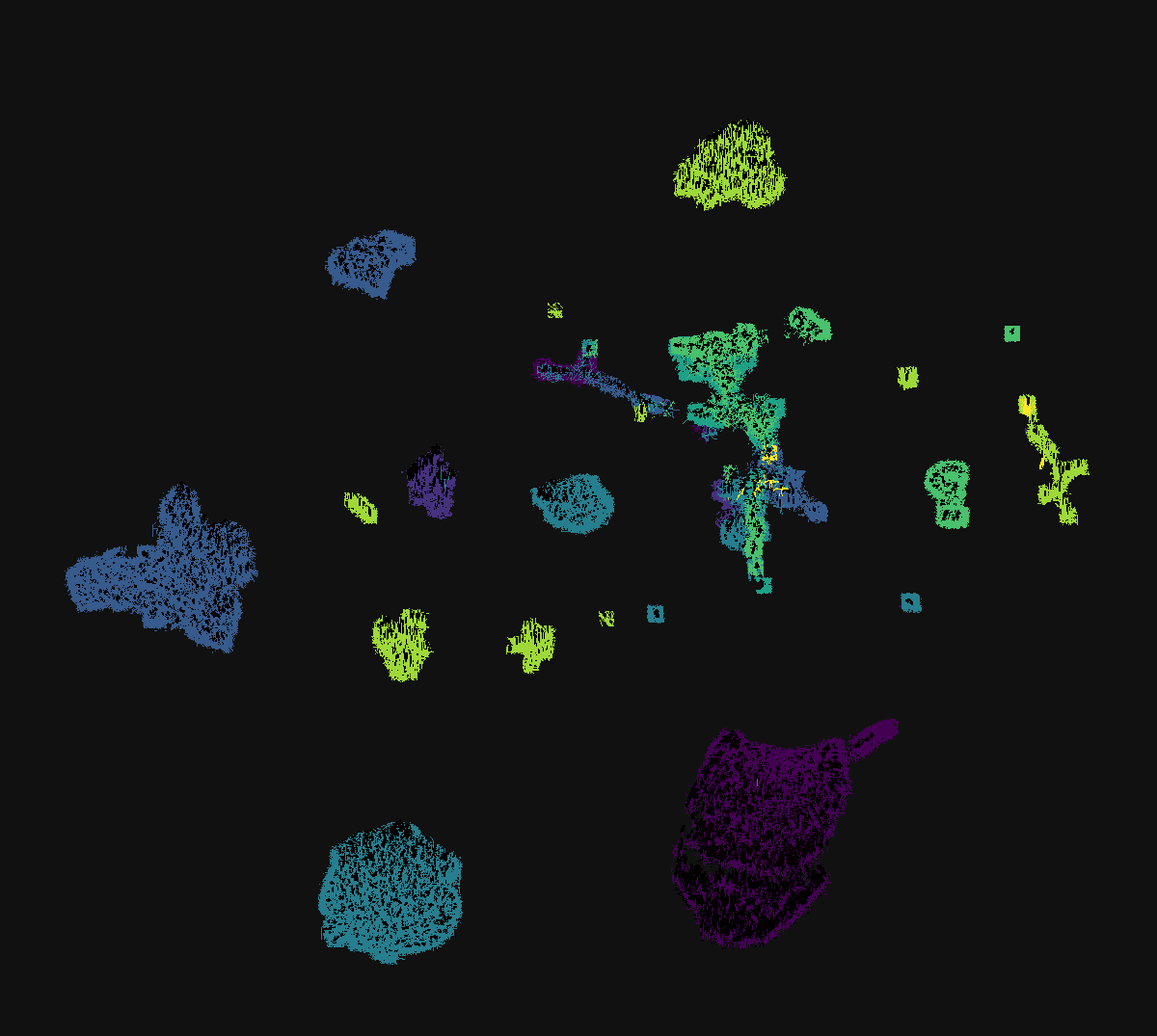}&
		\includegraphics[width=0.33\columnwidth]{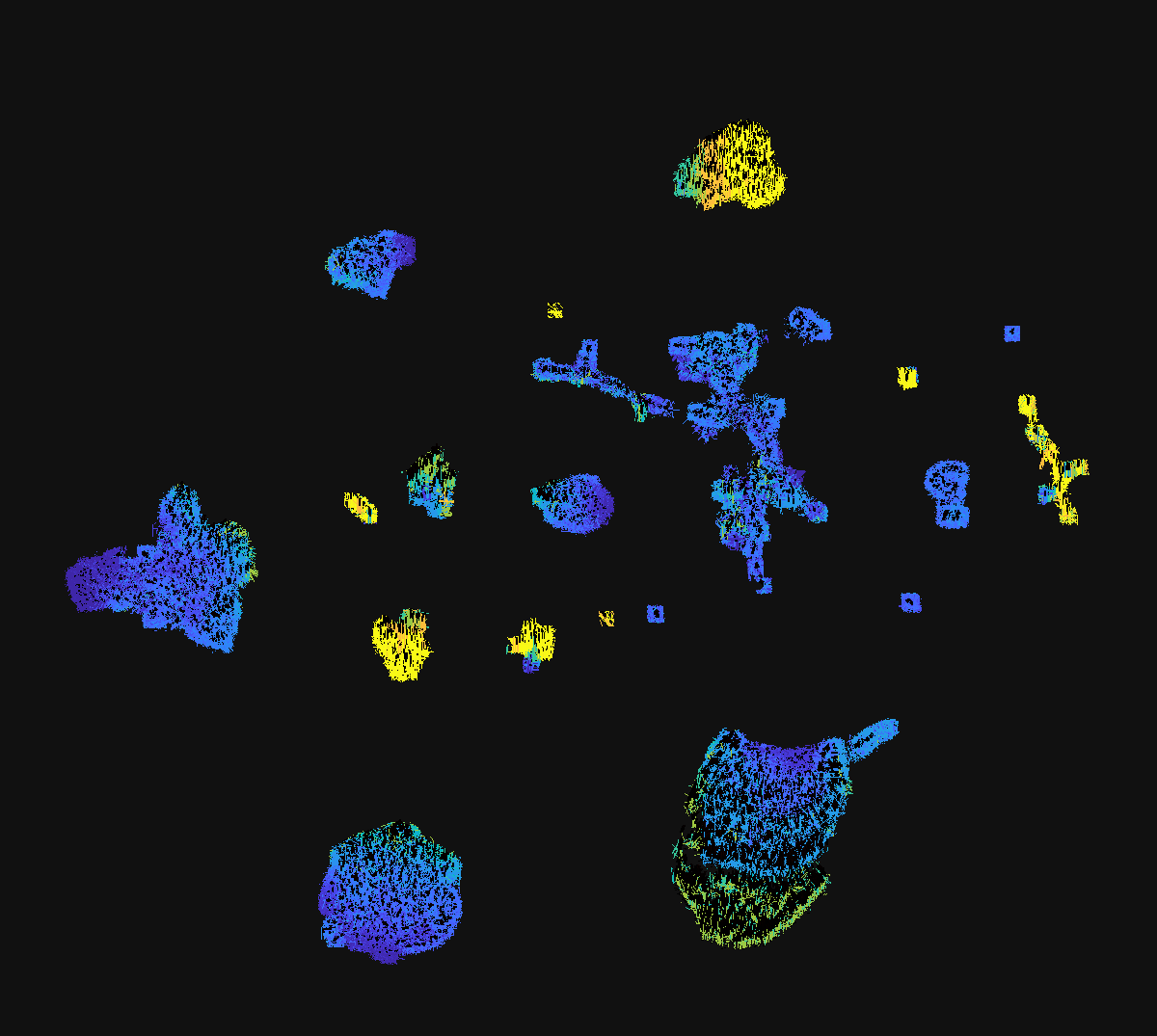}&
		\includegraphics[width=0.33\columnwidth]{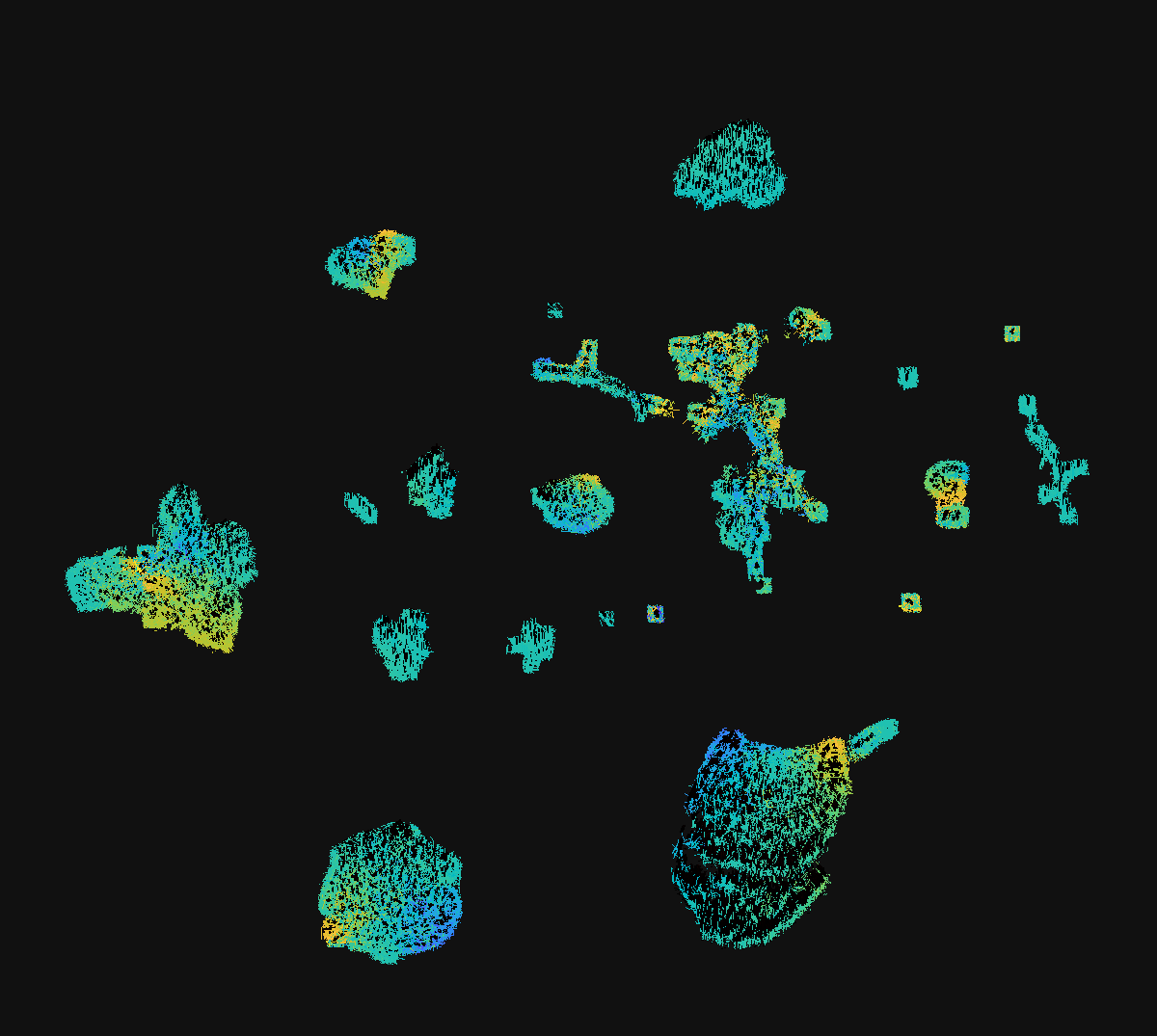}\\
		\includegraphics[width=0.33\columnwidth]{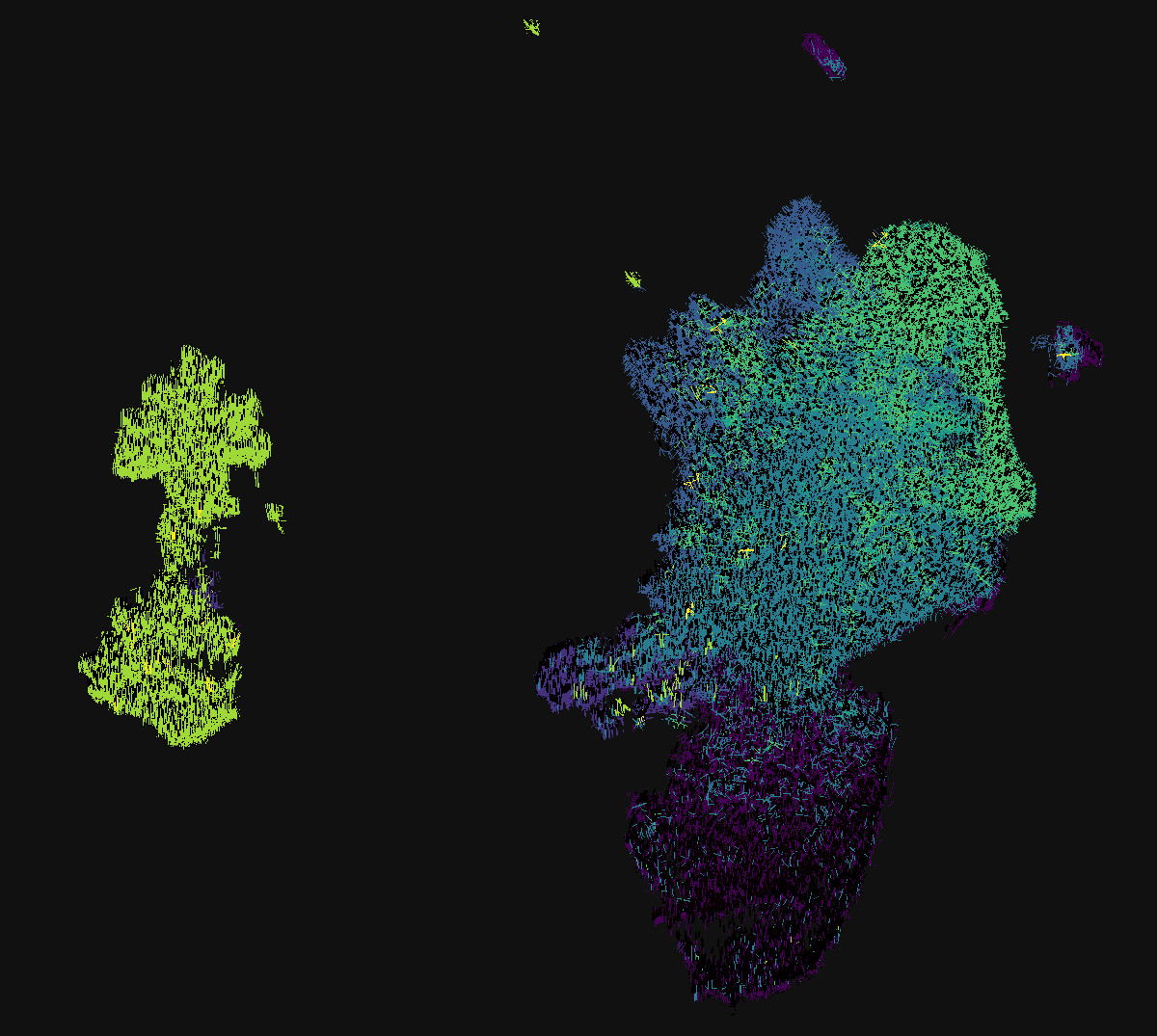}&
		\includegraphics[width=0.33\columnwidth]{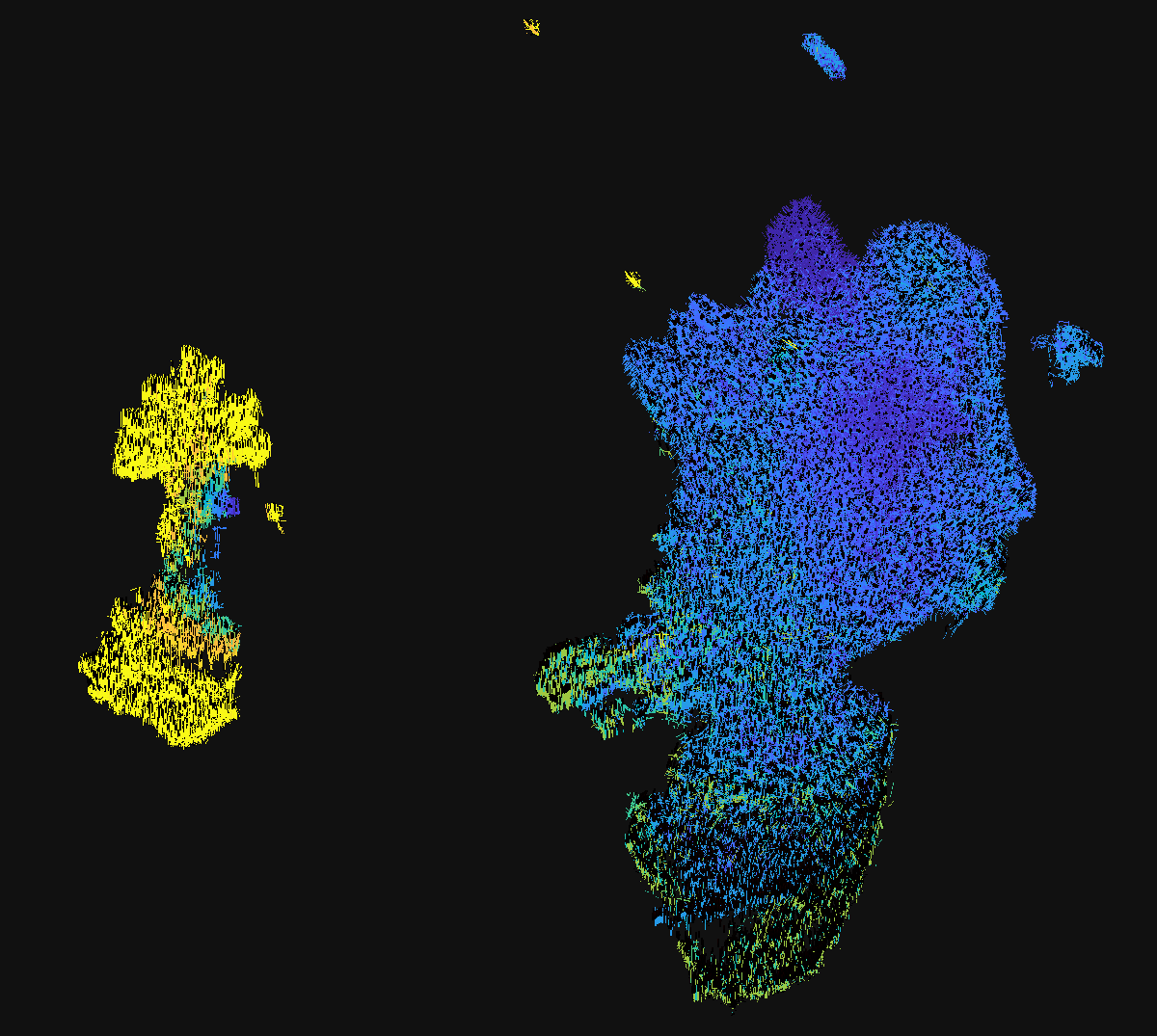}&
		\includegraphics[width=0.33\columnwidth]{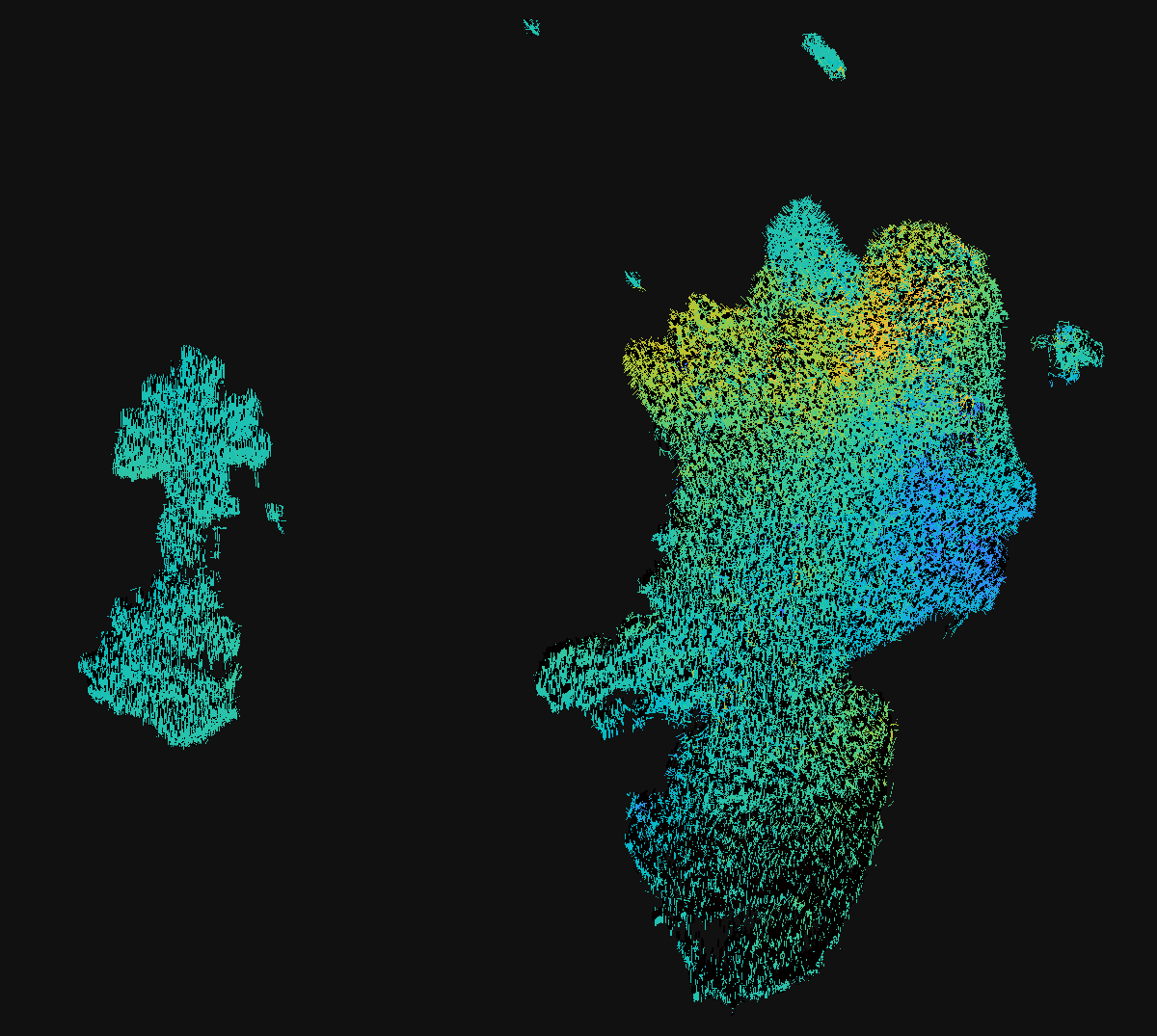}\\
	\end{tabular}
	\setlength\tabcolsep{6pt}
	\caption{UMAP visualizations of the latent spaces of the proposed setting (top) and with $\alpha_\mathrm{M}=0$ (bottom). \textit{Categories}: %
	\raisebox{-0.5ex}{\makebox[0pt][l]{\includegraphics[width=11.0ex,height=2.2ex]{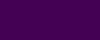}}} {\color{white}single-lane} , %
	\raisebox{-0.5ex}{\makebox[0pt][l]{\includegraphics[width=10.4ex,height=2.2ex]{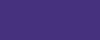}}} {\color{white}multi-lane} , \\%
	\raisebox{-0.5ex}{\makebox[0pt][l]{\includegraphics[width=11.5ex,height=2.2ex]{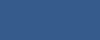}}} {\color{white}intersection} , %
	\raisebox{-0.5ex}{\makebox[0pt][l]{\includegraphics[width=16.8ex,height=2.2ex]{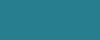}}} {\color{white}intersection-enter} , %
	\raisebox{-0.5ex}{\makebox[0pt][l]{\includegraphics[width=11.3ex,height=2.2ex]{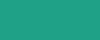}}} {\color{white}roundabout} , %
	\raisebox{-0.5ex}{\makebox[0pt][l]{\includegraphics[width=16.5ex,height=2.2ex]{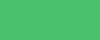}}} {\color{white}roundabout-enter} , %
	\raisebox{-0.5ex}{\makebox[0pt][l]{\includegraphics[width= 8.8ex,height=2.2ex]{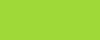}}} {\color{white}highway} and %
	\raisebox{-0.5ex}{\makebox[0pt][l]{\includegraphics[width=14.0ex,height=2.2ex]{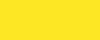}}} {\color{white}highway-enter} . %
	$\bar{v}$: \raisebox{-0.5ex}{\makebox[0pt][l]{\includegraphics[width=16.5ex,height=2.2ex]{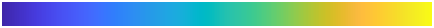}}}{\color{white}\,\unit[0]{m/s} $\dots$ \unit[42]{m/s}} . %
	$\bar{\psi}$: \raisebox{-0.5ex}{\makebox[0pt][l]{\includegraphics[width=12.5ex,height=2.2ex]{ref_n_n.png}}}{\color{white} \,$-\pi$ $\dots$ $+\pi$} .}
	\label{fig:vis}
\end{figure}

The resulting latent representations can be assessed by visualizing them. Therefore, UMAP is used to project the representation into two-dimensional space. In Fig. \ref{fig:vis}, the projections of two latent representations are depicted. The upper row shows the projection for the proposed setting, and the lower row shows the projection for the setting $\alpha_\mathrm{M}=0$ (turning off the metric learning). In the columns different color codings are used. In the first, the categories as in Sec \ref{sec:groups} are used. The second shows the average velocity of the trajectory and the last show values related to the orientation of the trajectory. The two latent representations were picked to demonstrate, how the latent representations differ, and how they can be analyzed using the visualization. It becomes clear, that the latent representation of the proposed setting provides well structure behavior in terms of categories, since the various infrastructure types are clearly separated. The model without the metric loss fails in separating 
the categories, instead two big clusters can be seen, one for the highway scenarios and one with all other scenarios. There is a strong relationship between the internal cluster structure and the shown features when using the proposed method. Therefore, the features  ($\bar{v}$, $\bar{\psi}$) show smooth course within the clusters (e.\,g. in the lower right cluster, the average speed increases from top to bottom). This is also true for the model without metric loss, but here in a more global scale. The analysis can support in understanding and validating the latent representations. Readers interested in exploring the projections in more detail may refer to the website published alongside the paper \url{https://jwthi.github.io/Expert-LaSTS/}. There, also the projections for the other settings as well as some alternative approaches are shown.

\subsection{Summary}
The different analysis perspectives highlight the performance with respect to specific tasks. Here, the overall performance is summarized and best model variants are discussed.

The only alternative approach able to perform considerably well in one of the perspectives is the classifier. However, when considering the other perspectives, the classifier does not seem to be a good choice either. All the other alternative approaches perform worse in all the perspectives, and hence are not appropriate for the presented problem setting.

Over all perspectives, three model variants should be highlighted. First, the proposed setting performs well throughout the perspectives. It seems to be the best selection when solving all tasks considerably well. The double size latent space setting is the second which performs well on different perspectives. With respect to feature stability it even outperforms the proposed setting. But, in terms of clustering it is worse. Therefore, if the focus is towards feature stability and less towards clustering this might be a good selection. The third and last model setting to be highlighted is using the LSTM encoder. With respect to detecting novel scenario types and clustering, it performs equally well as the proposed setting. However, it misses out slightly on the feature stability. 

\section{CONCLUSION}\label{sec:conc}
In this work, a method to design a latent space for traffic scenarios by means of expert-knowledge is presented. An automated mining strategy for traffic scenarios is introduced and used to find similar infrastructures and routes. This way, relative similarities as defined by expert objectives can be realized. The resulting latent space outperforms alternative approaches on various analysis perspectives, namely detecting novel scenario types, clustering and feature stability. The ablation study provides deep insight to the impact of various model parameters on the performance.

The method presented in this work can be used in the validation process for AVs. More precisely, it can support the analysis of scenarios as well as the detection of representative and novel scenarios.

Including further objects can be one possible direction for further research on the proposed method. Also, the performance when using real-world data can be analyzed in a next step.

\section{ACKNOWLEDGEMENT}
The authors acknowledge the support by the ZF Friedrichshafen AG.
Map data copyrighted OpenStreetMap contributors and available from \url{https://www.openstreetmap.org}.
\bibliographystyle{IEEEtran}
\bibliography{format,E:/IDA/Literatur/ref.bib}
\end{document}